\RequirePackage{fix-cm}
\documentclass[twocolumn]{svjour3}          
\smartqed  
\usepackage{graphicx}
\usepackage{amsmath}
\usepackage{multirow}
\usepackage{pifont}
\usepackage{graphicx}
\usepackage{amsmath}
\usepackage{amssymb}
\usepackage{array}
\usepackage{multirow, nicefrac}
\usepackage{pifont}
\usepackage{cite}
\usepackage{bbm}
\usepackage{arydshln}
\usepackage{xcolor}
\usepackage{tablefootnote}
\usepackage{threeparttable}
\definecolor{green}{rgb}{1,0,0}

\def\eg{\emph{e.g.}}
\def\ie{\emph{i.e.}}

\def\etal{\emph{et al.}}

\def\etal{{\em et al.\/}\, }

\def\0{{\bf 0}}
\def\1{{\bf 1}}

\def\bW{{\bf W}}
\def\bX{{\bf X}}

\def\st{\mbox{s.t. }}

\usepackage{listings,xcolor} 
\usepackage{tcolorbox}

\newcommand{\lstfont}[1]{\color{#1}\ttfamily}

\usepackage{xspace}
\usepackage[colorlinks,linkcolor=red,anchorcolor=blue,citecolor=purple,CJKbookmarks=True]{hyperref}

\newlength\savewidth\newcommand\shline{\noalign{\global\savewidth\arrayrulewidth
  \global\arrayrulewidth 1pt}\hline\noalign{\global\arrayrulewidth\savewidth}}
\def\eg{\emph{e.g.}}
\def\ie{\emph{i.e.}}
\def\etal{\emph{et al.}}

\usepackage{marvosym}

\begin{document}

\title{Mesa: A Memory-saving Training Framework for Transformers
}

\author{Zizheng Pan,
        Peng Chen,
        Haoyu He,
        Jing Liu,
        Jianfei Cai \\
        and Bohan Zhuang $\dagger$ 
}
\authorrunning{Z. Pan, P. Chen, H. He, J. Liu,  J. Cai, B. Zhuang} 

\institute{
Zizheng Pan,
        Peng Chen,
        Haoyu He,
        Jing Liu,
        Jianfei Cai 
        and Bohan Zhuang\at
              Faculty of Information Technology, Monash University \\
              \email{\{zizeng.pan, haoyu.he, jing.liu1, jianfei.cai, bohan.zhuang\}@monash.edu}, blueardour@gmail.com \\
        $\dagger$ Corresponding author.
}

\date{}

\maketitle

\begin{abstract}
   There has been an explosion of interest in designing high-performance Transformers.
   While Transformers have delivered significant performance improvements, training such networks is extremely memory intensive owing to storing all intermediate activations that are needed for gradient computation during backpropagation, especially for long sequences. 
   To this end, we present Mesa, a memory-saving training framework for Transformers.
   Specifically, Mesa uses exact activations during forward pass while storing a low-precision version of activations to reduce memory consumption during training. 
   The low-precision activations are then dequantized during back-propagation to compute gradients. Besides, to address the heterogeneous activation distributions in the multi-head self-attention layers, we propose a head-wise activation quantization strategy, which quantizes activations based on the statistics of each head to minimize the approximation error. To further boost training efficiency, we learn quantization parameters by running estimates. More importantly, by re-investing the saved memory in employing a larger batch size or scaling up model size, we may further improve the performance under constrained computational resources.
   Extensive experiments on ImageNet, CIFAR-100 and ADE20K demonstrate that Mesa can achieve flexible memory-savings (up to 50\%)
   during training while achieving comparable or even better performance. Code is available at \url{https://github.com/ziplab/Mesa}.
   
\end{abstract}

\section{Introduction}
\label{sec:intro}

Transformers have demonstrated stunning success in a wide range of natural language processing (NLP)~\cite{bert,albert} and computer vision (CV) tasks~\cite{pvt,detr,setr}. Inspired by the previous works on model scaling~\cite{resnet,efficientnet}, the recent researches on Transformers further push the performance forward with an increasing model size~\cite{gpt3,pvt,swin}.
However, training Transformer models requires a formidable amount of memory footprints, prohibiting common users with limited computational resources from doing related researches. For example, training a Swin-T~\cite{swin} with a batch size of 128 on 224$\times$224 images consumes at least 12G GPU memory, while Swin-B cannot fit into a 32GB V100 GPU under the same settings. Consequently, only a few parties can afford to train such large models. The huge memory consumption and the increasing resource inequalities make it difficult for the academic community to follow up on this area, based on the fact that most of the recent advanced Transformers are developed with industry participation. 

\begin{figure}[]
	\centering
	\includegraphics[width=\linewidth]{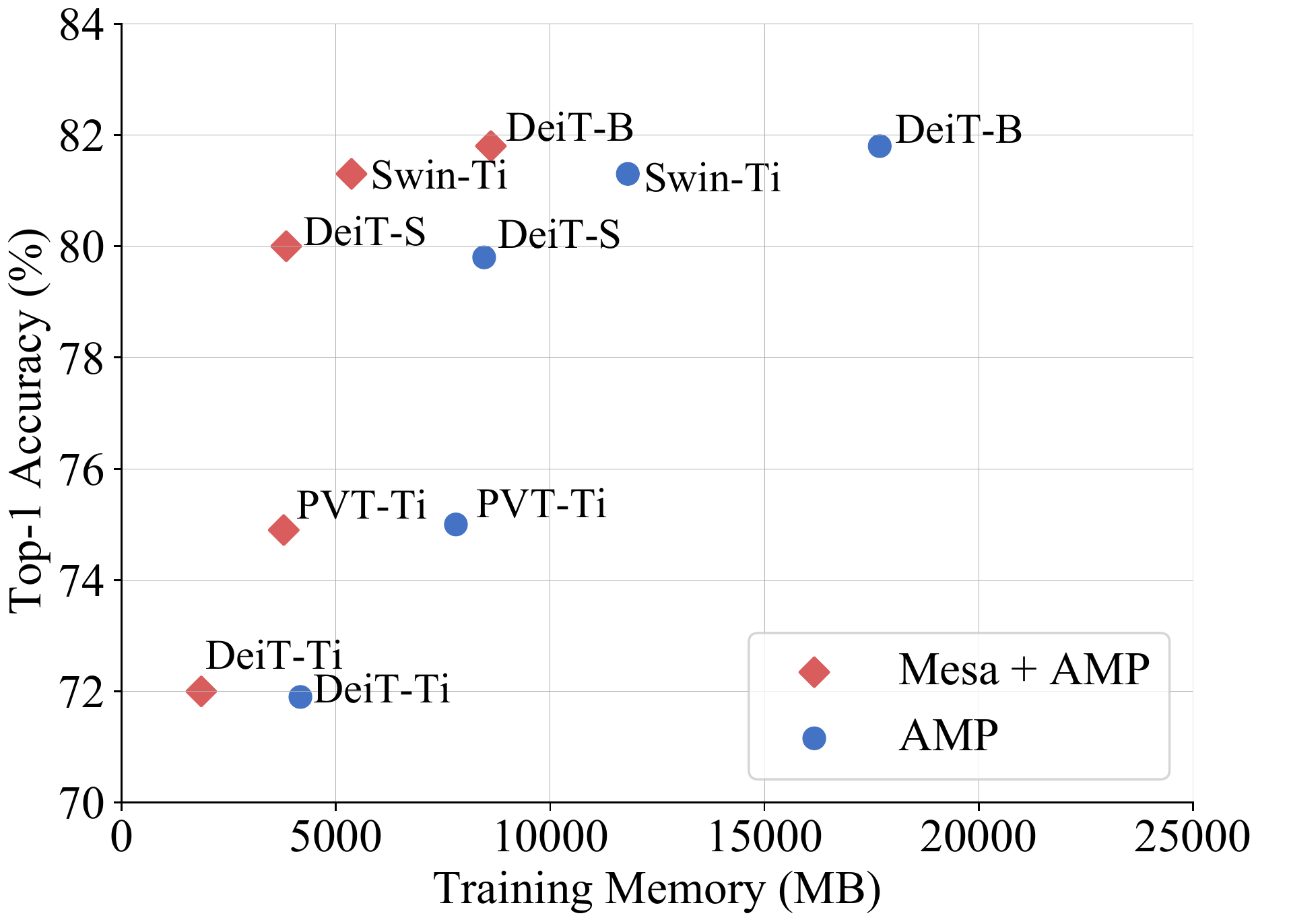}
	\caption{Comparisons of memory footprints during training with several state-of-the-art vision Transformers on ImageNet. 
	``AMP'' denotes the default automatic mixed-precision training~\cite{low_pre_train}. ``Mesa + AMP'' means that we train models with Mesa along with AMP. The memory footprint is measured with a batch size of 128 and an image resolution of 224$\times$224 on a single GPU. The proposed Mesa reduces around half of the memory consumption during training while achieving similar or even better performance compared to the default AMP training.}
	\vspace{-10pt}
	\label{fig:deit_compare}
\end{figure}

Fortunately, great efforts have been made to train deep neural networks with low memory. For example, existing representative techniques include sparsity~\cite{dyna_sparse_rep}, low-precision training~\cite{low_pre_train}, micro-batching~\cite{microbatching}, checkpointing~\cite{checkpointing} and gradient quantization~\cite{qsgd}. Among the existing memory-saving techniques, one promising direction is the activation compressed training (ACT)~\cite{backprop_mem}. Specifically, ACT stores low-precision approximate copies of activations at each layer while computing the forward pass exactly, which helps to reduce the overall memory consumption during training. The saved activations are then dequantized to the original precision in the backward pass to calculate gradients. This approach has been successfully applied to train ResNet~\cite{resnet} variants. However, existing works on ACT~\cite{backprop_mem,actnn,tinyscript} either focus on dedicated architectures or specifically target on convolutional neural works (CNNs). At present, there is no literature on compressing the commonly used operations in Transformers (\eg, Softmax, GELU~\cite{gelu} and LayerNorm~\cite{layernorm}). Moreover, no previous work considers the heterogeneously distributed activations in the multi-head self-attention (MSA) layers, especially the attentions. Consequently, we empirically observe that applying the naive activation bucketing strategy as in previous works~\cite{actnn} for the attention maps will make Transformers fail to converge. Therefore, none of the existing ACT frameworks can be directly applied to Transformer-based models. Targeting common operations in Transformer remains challenging and the effect is unknown.

To tackle the above challenges,  we present \textbf{Mesa}, a \textbf{Me}mory-\textbf{sa}ving 8-bit training framework for Transformers. Mesa covers all memory-hungry operations in Transformers, including matrix multiplications (MatMul), Softmax, LayerNorm and GELU.
Moreover, we propose a head-wise activation quantization strategy, which quantizes the activations based on each self-attention head. The motivation comes from two aspects. First, group-wise quantization, derived from product quantization~\cite{jegou2010product}, has shown to be effective in minimizing quantization error. The proposed head-wise quantization can be seen as a special case of group-wise quantization where the number of groups is equal to the number of self-attention heads. Second, previous studies have revealed that different self-attention heads in Transformers tend to learn different attention patterns\cite{attn_conv}. Empirically, we can also observe a large divergence of statistics among different heads as shown in Figure \ref{fig:act_dist}. Therefore, using shared quantization parameters across self-attention heads may result in highly degraded estimates of statistical quantities.

Besides, Mesa learns quantization parameters with efficient running estimates during training. Such approach brings additional benefits compared with gradient based approaches~\cite{choi2018pact,lsq_plus} or per-sample statistics~\cite{actnn} as it avoids extra computational and memory cost for learning quantization parameters.
In practice, we also observe that the running estimates performs favourably against the per-sample statistics in terms of both training speed and performance.

To the best of our knowledge, \textit{Mesa is the first ACT framework for Transformer-based models}.  It is also orthogonal to other memory saving techniques such as low-precision training~\cite{low_pre_train} and checkpointing~\cite{checkpointing}. Therefore, in practice, one can simultaneously apply automatic mixed-precision training (AMP) and checkpointing along with Mesa to achieve more memory-saving. 
As a bi-product of significant memory reduction during training, we can use a larger batch size and/or train a larger Transformer model to enable fully-utilization of available GPU cores. 
Furthermore, we are able to re-invest the saved memory during training by constructing $3.3\times$ deeper or $2.2\times$ wider DeiT-B, or training DeiT-B with $1.5\times$ larger image resolution.
Note that as in common practice~\cite{actnn,checkpointing,DBLP:conf/stoc/IndykM98}, there is a trade-off between the memory cost and training speed for Mesa. In this case, Mesa is designed to flexibly compress different operations/modules to achieve a target memory-saving with acceptable training overhead. Moreover, for a large number of researchers with only entry/middle-level GPUs, their first thing is that Transformer can be run in their machines even with small-size models, before considering throughput. To this end, what Mesa offers to the community is a comprehensive study of customizing ACT to Transformers, along with an efficient and effective plug-and-play module with CUDA implementation that can be easily used in a variety of Transformer models for flexible memory-saving.
In a nutshell, we summarize our contributions as follows: 
\begin{enumerate}
\setlength{\itemsep}{1.5ex}  
\setlength{\itemindent}{1.1em}  
\item We propose a memory-saving 8-bit training framework for Transformers, namely Mesa. Mesa is implemented with a fast CUDA kernel and can be easily adapted to any Transformer projects.

\item We observe the heterogeneously distributed activations in self-attention heads. For this, we propose a head-wise activation quantization strategy to minimize the approximation error in MSA layers. Besides, we use running estimates to learn quantization parameters, which performs well with negligible additional cost.

\item Extensive experiments on ImageNet, CIFAR-100 and ADE20K have demonstrated that Mesa can reduce $\sim$50\% memory footprint during training with comparable or even better performance than the standard mixed-precision training scheme.
\end{enumerate}
\section{Related Work}
\label{related_work}

\subsection{Transformers}
Transformer is initially proposed by Vaswani~\etal~\cite{transformer} for machine translation. A standard Transformer consists of an embedding layer, several Transformer blocks and a task-specific head, where each block contains an MSA layer and a position-wise feed-forward (FFN) layer. Later on, Transformer has been extended into a wide range of tasks. In the area of computer vision, vision Transformers (ViTs) have attracted great attentions recently.
For example, Dosovitskiy~\etal~\cite{vit} proposed a standard Transformer architecture for image recognition, which achieved competitive results on ImageNet compared to CNNs. Subsequent works have improved ViTs from different aspects, such as incorporating pyramid features~\cite{pvt,swin,hvt} , adopting convolutional layers to enhance the model locality~\cite{t2t}, or exploring a well performed architecture with neural architecture search (NAS)~\cite{glit,autoformer}.
However, to train a Transformer usually requires intensive computational resources. For example, the typical setting~\cite{deit} to train a ViT on ImageNet requires a batch size of 1,024 on 8 NVIDIA V100 GPUs. As a result, only a few parties are capable of running such experiments.
Besides, it also makes it difficult for researchers to explore a larger design space for Transformer architectures. To address this problem, we propose to reduce the  memory cost of Transformers during training by 8-bit activation compressed training, making the experiments affordable.

\subsection{Quantized Training}
Quantized training aims to improve the model efficiency at training time or inference time by quantizing model weights, activations or gradients into low-precision values. Existing methods can be roughly categorised into two folds: 1) quantizing a pretrained model with or without retraining~\cite{lsq,qil,ptq_wang,data_free_quant}, and 2) training a quantized model from scratch~\cite{xnor_net,binaryconnect,low_pre_train,qsgd}. In Transformers, the majority of the literature belongs to the first category. For example, 8-bit~\cite{q8bert,qbert} or even lower-bits~\cite{binary_bert} quantization has been proposed to speed up the inference. In contrast, this paper focuses on training Transformers from scratch. Different from previous works~\cite{q8bert,ternary_bert} that carry out low-precision computations during either the forward pass or the backward pass, we store the approximated low-precision activations for memory saving during training while still computing the forward pass exactly. Therefore, we do not change the forward pass behavior of models.

\subsection{Memory-efficient Training}
Low-memory training is appealing as it enables large-scale model training in resource-constraint scenarios. A plethora of methods have been proposed in this area. For example, Mostafa~\etal~\cite{dyna_sparse_rep} proposed to reduce the model and optimizer memory by dynamic sparse reparameterization. Micikevicius~\etal~\cite{low_pre_train} introduced mixed precision training, which utilizes mixed half precision (16-bits) and full precision (32-bits) for training. Huang~\etal~\cite{microbatching} proposed to split the mini-batch into smaller subsets and then accumulate the gradients until the entire minibatch has been processed. Besides, gradient checkpointing~\cite{checkpointing} is also a common practice to reduce the activation memory.

Orthogonal to the above approaches, activation compressed training (ACT)~\cite{backprop_mem} has recently been proposed to reduce the storage of activations that are required for gradient computation.
This method was first introduced by Chakrabarti~\etal~\cite{backprop_mem} for ResNet training. Subsequent works such as TinyScript~\cite{tinyscript} and ActNN\cite{actnn} extend this framework by introducing non-uniform quantization and mixed-precision quantization, respectively. However, all these methods specifically target on CNNs and do not consider the unique components of Transformers, such as LayerNorm, Softmax and GELU. Moreover, the quantization scheme for them is problematic for Transformers due to the heterogeneous activation distributions in an MSA layer. Consequently, none of the existing methods can be directly applied to Transformer based models. 
In this paper, we customize the ACT framework to significantly reduce the resource requirement of training Transformers while keeping their outstanding performance. 
\section{Method} \label{sec:method}
In this section, we first describe the overall framework of Mesa. Then we introduce our proposed head-wise activation quantization and the strategy for learning quantization parameters. Lastly, we provide the details of the system implementation and discuss the overhead of Mesa.
\begin{figure}[]
	\centering
	\includegraphics[width=\linewidth]{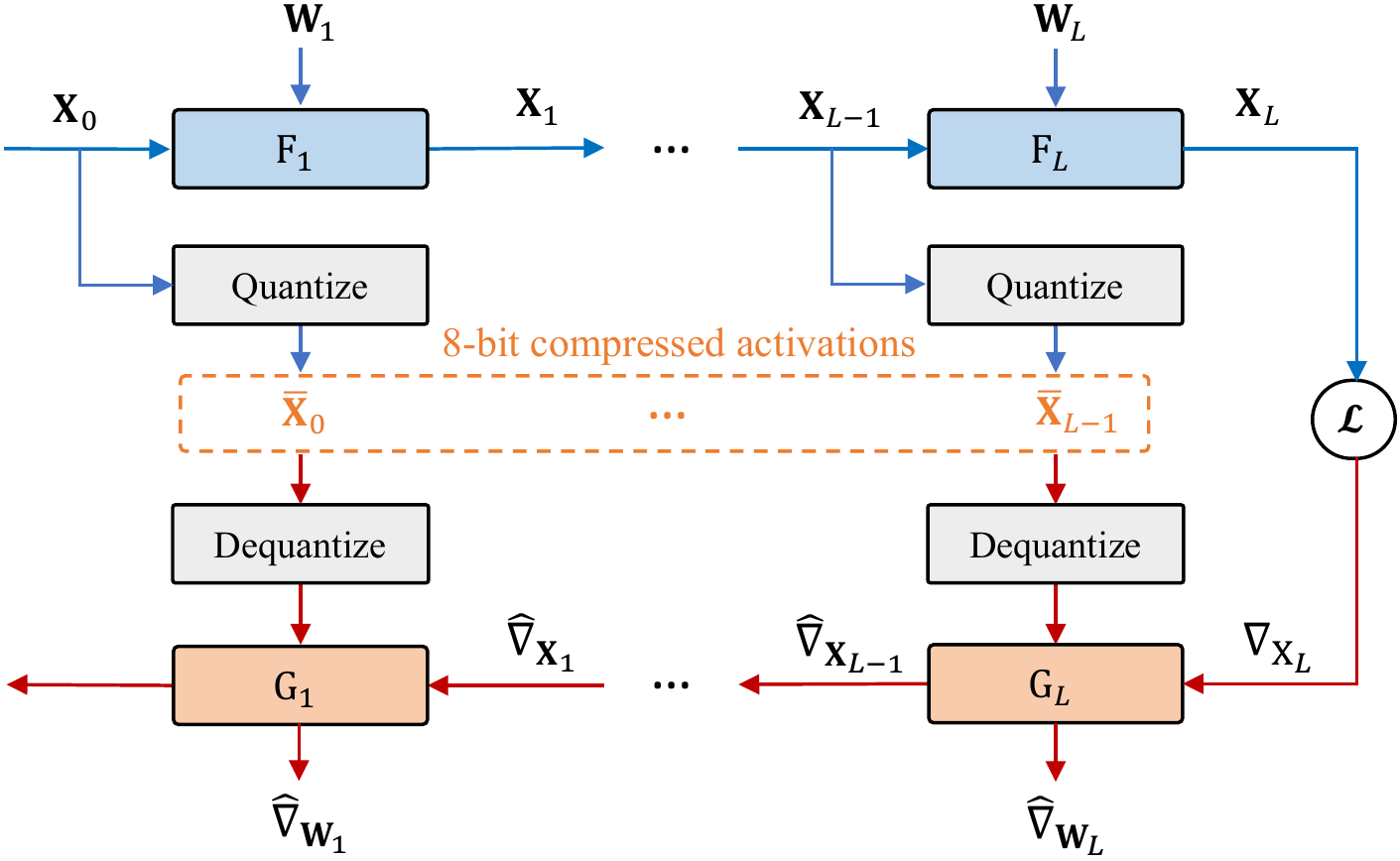}
	\caption{The pipeline of the proposed Mesa for training Transformers, where we compress the activations into low-precision values during training to achieve memory reduction while still propagating the exact activations during the forward pass. The dequantized activations are used to compute gradients during backpropagation. 
	Blue and red lines represent the forward and backward passes, respectively}
	\label{fig:framework}
\end{figure}

\subsection{Overview of Mesa} \label{sec:overview}
To reduce the memory consumption of Transformers at training time, we introduce Mesa, a generic memory-saving training framework for Transformers.
The overall pipeline of Mesa is depicted in Figure~\ref{fig:framework}. In general, Mesa saves low-precision approximated activations during training for backpropagation while still using exact activations for the forward pass. Specifically, denoting $\bX_{l-1}$ as the input to the $l$-th layer in a Transformer, the output of the $l$-th layer can be formulated by
\begin{equation}
    \bX_l = \mathrm{F}_l(\bW_l, \bX_{l-1}),
\end{equation}
where $\mathrm{F}_l$ and $\bW_l$ represent the function and learnable parameters of the $l$-th layer, respectively.

\begin{figure*}[]
	\centering
	\includegraphics[width=0.8\linewidth]{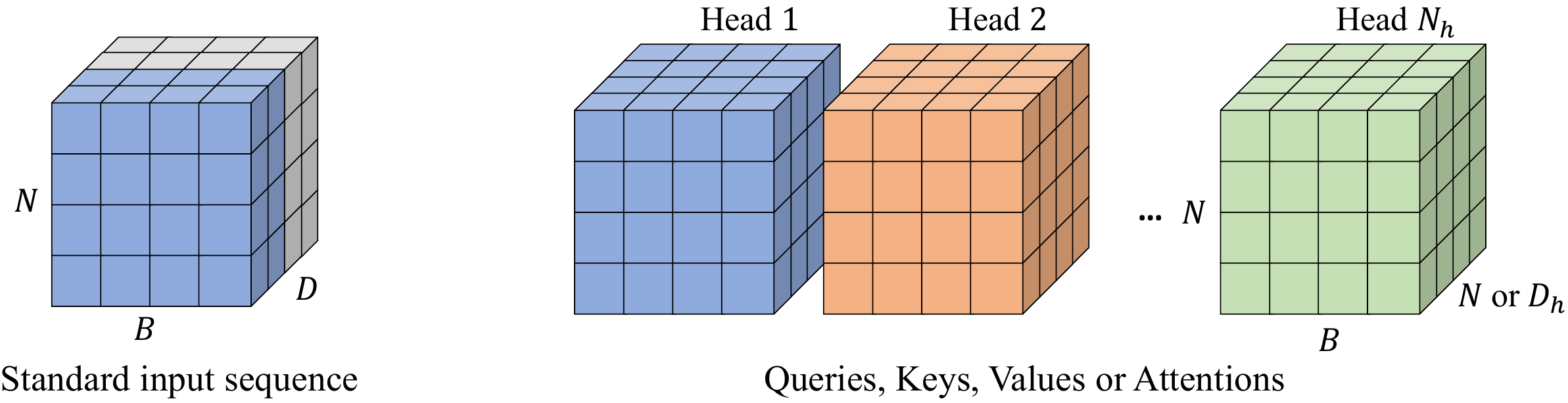}
	\caption{The different group-wise bucketing strategies in Mesa. 
	Different colors represent different quantization groups.
	Let $B$, $N$ and $D$ be the batch size, the sequence length and the number of hidden channel dimensions, respectively. $N_h$ and $D_h$ refer to the number of self-attention heads and the head dimensions, respectively. For a standard input sequence, we group the activations based on a certain number of hidden channels ($D/2$ in this example).
	For the queries, keys, values and attentions, we group the activations based on self-attention heads. Best viewed in color.}
	\vspace{-10pt}
	\label{fig:bucketing}
\end{figure*}

In a standard training, the input $\bX_{l-1}$ is saved in the GPU memory in order to calculate gradients during backpropagation, where the saved activations at all layers take up the majority of the memory consumption during training, especially when equipped with a large batch size.
To reduce the memory footprint, we propose to only save the compressed activations $\Bar{\bX}_{l-1}$ instead of the full-precision counterparts $\bX_{l-1}$ during the forward pass. Such compression is achieved by quantization, which quantizes the exact activations into low-precision values.
During backpropagation, 
$\Bar{\bX}_{l-1}$ is dequantized into the original precision values $\hat{\bX}_{l-1}$ for gradient calculation. In this way, the gradients at the $l$-th layer can be approximated by
\begin{equation} \label{eq:appox_grad}
    \hat{\nabla}_{\bX_{l-1}}, \hat{\nabla}_{\bW_l} = \mathrm{G}_l(\hat{\nabla}_{\bX_l}, \hat{\bX}_{l-1}, \bW_l),
\end{equation}
where $\hat{\nabla}_{\bX_{l-1}}, \hat{\nabla}_{\bW_l}$ represent the approximated gradients for the input $\bX_{l-1}$ and the learnable parameters $\bW_l$, and $\mathrm{G}_l$ is the gradient function at the $l$-th layer.

It is worth noting that this strategy has little effect on the training performance, since it only introduces modest approximation errors to the natural gradient noise from distributed training and stochastic gradient descent (SGD)~\cite{DeanCMCDLMRSTYN12,terngrad}.
In the next section, we introduce a head-wise activation quantization scheme to further improve the fidelity of gradients during training.

\subsection{Head-wise Activation Quantization} \label{sec:head_wise_quant}
As shown in \cite{jegou2010product}, a fine quantization granularity is favorable to minimize the approximation error. In practice, layer-wise quantization is fast but may introduce a large quantization error, while channel-wise quantization can be more accurate but comes with extra computational and memory cost. In light of this, group-wise quantization balances the two sides and has been widely adopted in the literature~\cite{qbert,q8bert}. 

\begin{figure}[]
	\centering
	\includegraphics[width=1.0\linewidth]{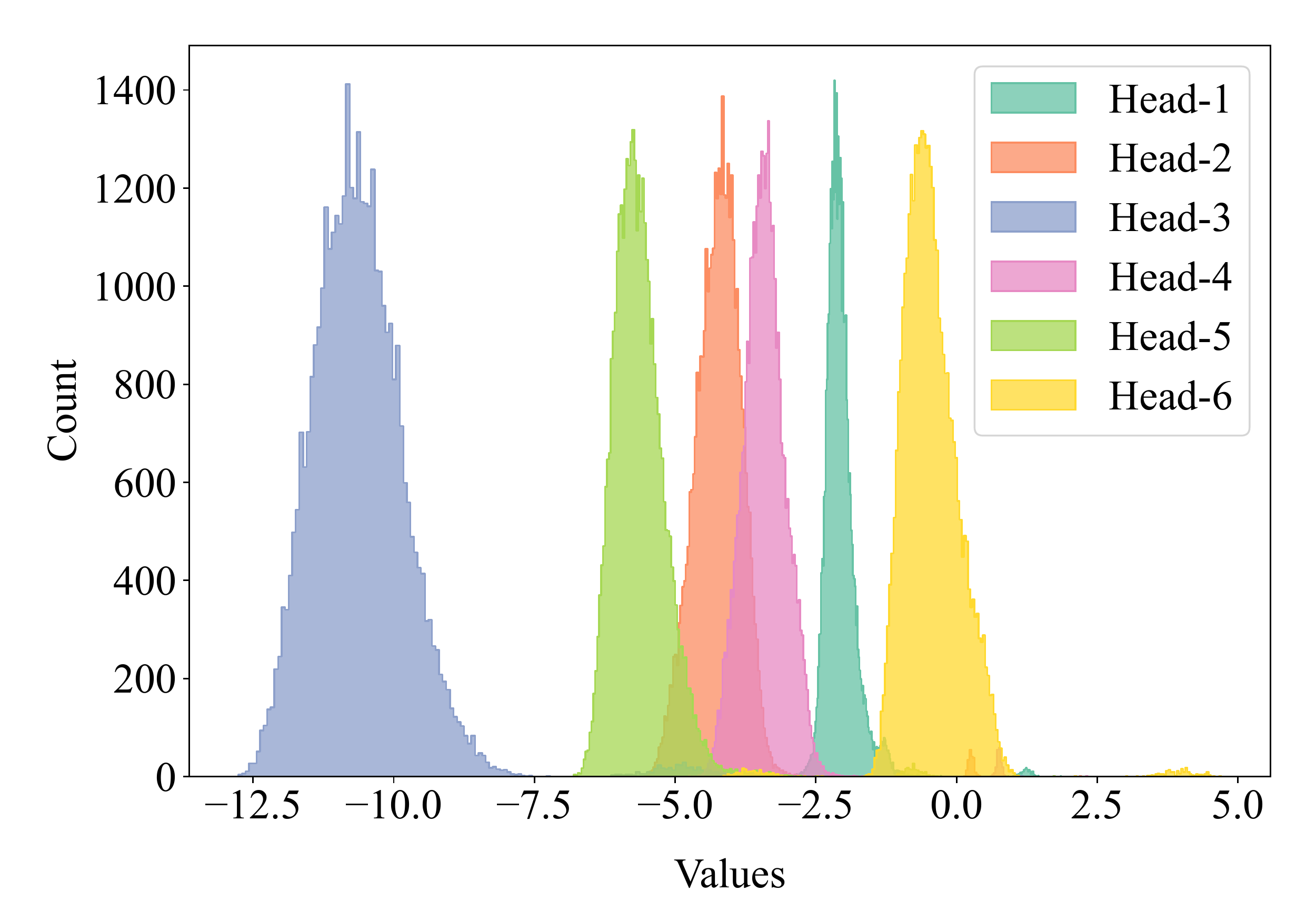}
	\caption{Activation distributions before the Softmax layer at the 11-th block of DeiT-S~\cite{deit}. We slice the activation tensor into different groups 
	based on self-attention heads
	and visualize each group in different colors. Best viewed in color.}
	\label{fig:act_dist}
\end{figure}

A naive grouping strategy in the previous works~\cite{actnn,tinyscript} is to slice a tensor into fixed-size buckets regardless of the tensor dimensions, which is however problematic for Transformers as it ignores the fact that different self-attention heads usually have quite different attention patterns~\cite{attn_conv}, \ie, the activation distributions across heads are with distinct means and variances in an MSA layer.
In Figure~\ref{fig:act_dist}, we visualize the activations before the Softmax layer at the 11-th block of DeiT-S. Clearly, the activation at each self-attention head should have its unique clipping range and offset. Such phenomenon can be observed for the queries, keys, values and attentions across all Transformer blocks. 

\noindent\textbf{Quantization scheme.}
To address the heterogeneously distributed activations in self-attention heads, we propose a head-wise activation quantization strategy.
Specifically, with $x \in \bX^h$ being one element in the input activations at the $h$-th head in an MSA layer, we quantize the activations to 8-bit by 
\begin{equation} \label{eq:quant}
    \Bar{x} = \mathrm{clip}(\mathrm{round}((x - \beta^h) \cdot \frac{255}{\alpha^h}), 0, 255 ),
\end{equation}
where $\alpha^h$ and $\beta^h$ are learnable parameters representing the quantization clipping range and the offset at the $h$-th head, respectively. $\mathrm{clip}(x, x_{low}, x_{up})$ clips any number $x$ into the range of $[x_{low}, x_{up}]$. $\mathrm{round}(\cdot)$ denotes the rounding function.
Here we adopt the stochastic rounding ~\cite{binaryconnect} as it theoretically guarantees smaller probabilistic error bounds~\cite{croci2021stochastic} compared to the nearest rounding.
Specifically, it can be formulated as 
\begin{equation}
    \mathrm{round}(x) = 
    \begin{cases}
    \lceil x \rceil & \text{with probability } p=x - \lfloor x \rfloor,\\
    \lfloor x \rfloor              & \text{otherwise}.
\end{cases}
\end{equation}
During backpropagation, we dequantize the activations at this layer into the original precision by
\begin{equation} \label{eq:dequant}
    \hat{x} = \Bar{x} \cdot \frac{\alpha^h}{255} + \beta^h.
\end{equation}
The dequantized activations are then used to calculate gradients as in Eq.~(\ref{eq:appox_grad}). In Section~\ref{sec:ablation}, we will show the effectiveness of the proposed quantization scheme by comparing it to other strategies, such as symmetric quantization and nearest rounding.

\noindent\textbf{Learning quantization parameters with running estimates.}
To calculate the quantization parameters $\alpha$ and $\beta$ at each layer, some previous works propose to use per-sample statistics~\cite{actnn,tinyscript} or gradient-based approaches~\cite{choi2018pact,lsq_plus}. Specifically, the methods of per-sample statistics utilise the current min-max values of each sample at each layer to calculate quantization parameters, which is inefficient and consumes additional memory for storing the quantization parameters.
For gradient-based approaches, asymmetric quantization methods such as LSQ+~\cite{lsq_plus} may increase the memory footprint as they need to save both the compressed activations and a large amount of context (\eg, the rounding errors for learning $\alpha$) that are needed to calculate gradients during backpropagation, contradicting our aim of memory saving. 
Alternatively, symmetric quantization approaches such as PACT~\cite{choi2018pact} do not have this problem as they only require binary values to calculate the gradients for $\alpha$. However, we will show in Section~\ref{sec:asym_sym} that symmetric quantization achieves poor performance on Transformers compared to both baselines and asymmetric quantization.

To this end,
we propose to utilise running estimates\cite{JacobKCZTHAK18} to update the quantization parameters during training, which can be expressed as 
\begin{equation} \label{eq:ema_alpha}
    \alpha^h = \lambda \alpha^h + (1 - \lambda)(\mathrm{max}(\bX^h) - \mathrm{min}(\bX^h)),
\end{equation}
\begin{equation} \label{eq:ema_beta}
    \beta^h = \lambda \beta^h + (1 - \lambda)\mathrm{min}(\bX^h),
\end{equation}
where $\lambda$ is a hyperparameter. As the quantization parameters are shared within each head across different samples, using running estimates can save more memory at training time.
For example, letting $N_h$ be the number of self-attention heads in an MSA layer and $B$ be the batch size, with running estimates, we only need to store $2N_h$ additional parameters for the Softmax layer, which is negligible compared to the overall memory footprint. In contrast, using per-sample statistics requires to save $2BN_h$ additional parameters for one layer.
Furthermore, we will show in Section~\ref{sec:ablation} that using running estimates can achieve faster throughput at training time.

\noindent\textbf{Other types of activations.}
It is worth noting that we adopt the head-wise bucketing strategy for queries, keys, values and attentions in a Transformer model. For other types of activations such as the standard input sequence in FFN layers, we apply the channel group-wise quantization scheme where multiple channel dimensions of activations are grouped. Figure~\ref{fig:bucketing} summarizes the different bucketing strategies of Mesa.

\subsection{System Implementation} \label{sec:sys_implementation}
Mesa is built upon the popular PyTorch~\cite{pytorch} framework. It is a standalone package that can be directly adopted into any Transformer projects. To further accelerate the quantization procedure (\ie, Eq.~(\ref{eq:quant}) and Eq.~(\ref{eq:dequant})) during training, we implement a fast CUDA kernel to improve the training efficiency. Overall, Mesa is able to compress all memory-hungry operations in a Transformer, including MatMul, LayerNorm, Softmax and GELU. Moreover, to support downstream tasks such as semantic segmentation, we also cover commonly used layers in CNNs, such as the convolutional layer and ReLU. For these layers, we adopt the standard group-wise quantization strategy as mentioned in Section~\ref{sec:head_wise_quant} where we group the convolutional activations based on a fixed number of channels. By default, we set the number of quantization groups to be the number of heads at a Transformer block.

\section{Discussions} \label{sec:discussion}
\subsection{The Speed and Memory Trade-off} While both the training speed and the memory consumption are critical bottlenecks for Transformer training, this work focuses on the latter. As in common practice (\eg, checkpointing~\cite{checkpointing} and hashing~\cite{DBLP:conf/stoc/IndykM98}), there is a trade-off between the speed and memory consumption at training time for all ACT frameworks~\cite{actnn,tinyscript}.
In Mesa, the additional training overhead includes the quantization/de-quantization, the
computation of min/max values, stochastic rounding and tensor reshaping for highly accurate CUDA dispatching. These operations can happen in almost all layers to achieve significant memory-saving while sacrificing the training speed.
In an extreme case, Mesa could result in the throughput decreases by half on a single GPU when compressing a whole model.

On the other hand, considering that distributed training has been widely adopted in modern deep learning~\cite{bert,vit,pvt}, the training cost can be dominated by data loading and communication overhead among GPUs, especially for small models.
For example, we will show in Section~\ref{sec:main_res} that training PVT-Ti~\cite{pvt} with Mesa only requires additional 15\% GPU hours. Besides, to further mitigate the additional training overhead, we modularize Mesa such that it can flexibly target different components in a model during training. In practice, we suggest to compress the most memory-hungry modules (\eg, backbones) to achieve a target memory-saving with acceptable training speed.
In Section~\ref{sec:ablation}, we show in Table~\ref{tab:compress_layer_types} that compressing different operations (\eg, Softmax and GELU) can achieve different speed and memory trade-offs. We will also provide the experiments on compressing different modules (\eg, MSA and FFN) in Table~\ref{tab:compress_msa_ffn} for reference.

\subsection{Mixed-precision and Lower Bits Quantization}
Apart from the 8-bit quantization used in Mesa, mixed-precision quantization~\cite{haq} could potentially bring more benefits. However, it will also introduce extra training overhead as it has to calculate the quantization bits for each layer, thus again slowing down the training speed.
Furthermore, compared to 8-bit integer, the computation and storage of lower bit values have not been well supported in existing CPUs/GPUs and deep learning frameworks (\eg, PyTorch).
For this reason, we choose the fixed 8-bit quantization in Mesa for a better trade-off between training speed and model performance.

\subsection{Relation to Other Low-memory Training Techniques}
As mentioned in Section~\ref{related_work}, Mesa is orthogonal to checkpointing~\cite{checkpointing}, micro-batching~\cite{microbatching} and mixed-precision training~\cite{low_pre_train}. Therefore, Mesa can be applied along with these techniques for more memory reduction, as shown in Table~\ref{tab:compare_checkpoint_ga} and Figure~\ref{fig:deit_compare}.
Besides, checkpointing targets the computational graph and keeps performance constant. In practice, one has to manually set the checkpointing positions, which is not optimal for the memory-speed trade-off. 
In contrast, Mesa is more flexible for compressing specified operations to match a target training memory, as shown in Table~\ref{tab:compress_layer_types}. Mesa can also potentially improve performance as well as easily identifying the most memory-intensive operations/modules for training. For micro-batching~\cite{microbatching}, it imitates a larger batch size but may affect model performance. Moreover, training with different batch sizes makes it difficult for a fair comparison with related works.
For the existing ACT works, they either cannot apply to Transformers~\cite{backprop_mem,tinyscript} or fully apply to Transformers~\cite{actnn}, and thus they are not directly comparable. One closely related work is ActNN~\cite{actnn}, for which we show in Section~\ref{sec:compare_actnn} that Mesa achieves better memory-saving and speed trade-off. 
\begin{figure*}[]
	\centering
	\includegraphics[width=\linewidth]{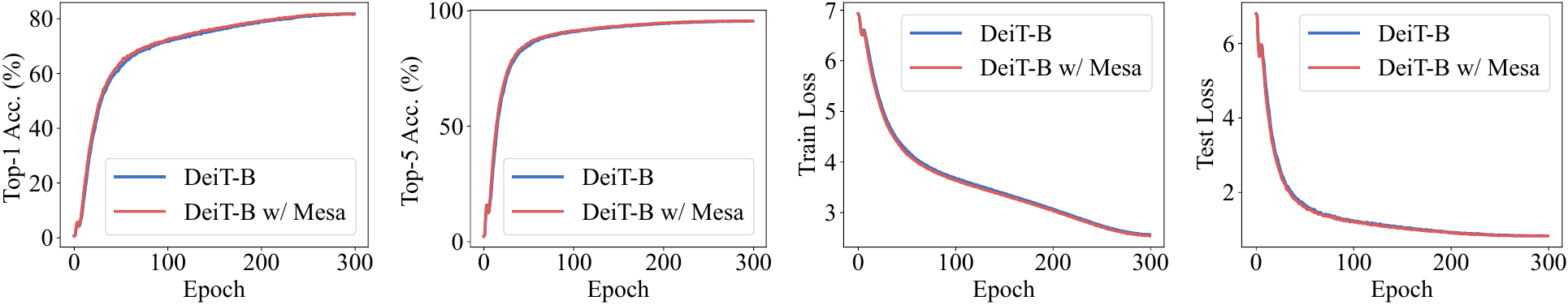}
	\caption{Training curves comparison for DeiT-B with/without Mesa on ImageNet.}
	\label{fig:deit_b_training_curve}
\end{figure*}

\begin{table*}[]
\centering
\caption{Classification results on ImageNet. ``*'' denotes our retrained baseline. The training time memory footprint is measured with a batch size of 128. All models use the default mixed-precision training~\cite{low_pre_train} as in baselines~\cite{deit,pvt,swin}. We measure the total training time by GPU hours w.r.t. a single NVIDIA V100 GPU.  Note that for inference, Mesa does not increase the model parameter (1st column) and does not affect the computational cost measured in FLOPs (2nd column).}
\renewcommand\arraystretch{1.2}
\scalebox{1.0}{
\begin{tabular}{l|ccccc}
Method         & Param (M) & FLOPs (G) & Train Memory (MB) & GPU Hours & Top-1@Acc. (\%) \\ \shline
DeiT-Ti*~\cite{deit}        & 5         & 1.3       & 4,171              & 440           & 71.9            \\
\textbf{DeiT-Ti w/ Mesa} & 5         & 1.3       & \textbf{1,858}              & 540          & \textbf{72.1}            \\ \hline
DeiT-S         & 22        & 4.6       & 8,459             & 520           & 79.8            \\
\textbf{DeiT-S w/ Mesa}  & 22        & 4.6       & \textbf{3,840}              & 620           & \textbf{80.0}            \\ \hline
DeiT-B         & 86        & 17.5      & 17,691             & 600           & 81.8            \\
\textbf{DeiT-B w/ Mesa}  & 86        & 17.5      & \textbf{8,616}              & 1,170           & \textbf{81.8}    \\  \hline
Swin-Ti~\cite{swin}         & 29        & 4.5      &  11,812                   &  480           & 81.3            \\ 
\textbf{Swin-Ti w/ Mesa}  & 29        & 4.5      & \textbf{5,371}                  & 820           &  \textbf{81.3}    \\ \hline
PVT-Ti~\cite{pvt}         & 13        & 1.9      &  7,800                   & 520             & 75.1            \\ 
\textbf{PVT-Ti w/ Mesa}  & 13        & 1.9      & \textbf{3,782}                  & 600           &  74.9     
\end{tabular}
}
\label{tab:classification}
\end{table*}

\section{Experiments}
\label{experiment}

\subsection{Main Results} \label{sec:main_res}
\noindent\textbf{Dataset and evaluation metrics.}
We conduct experiments on the ImageNet (ILSVRC2012)~\cite{imagenet} dataset, which contains $\sim$1.2M training images from 1K categories and 50K validation images. Following the common practice~\cite{deit,vit}, we measure the model performance by Top-1 accuracy. In addition, we also report the memory consumption at training time and the overall GPU hours for training each model.

\noindent\textbf{Compared methods.}
To demonstrate the effectiveness of Mesa, we evaluate our framework on several state-of-the-art vision Transformers, including DeiT~\cite{deit}, Swin~\cite{swin} and PVT~\cite{pvt}. DeiT is a standard vision Transformer which inherits the similar architecture from the original Transformer~\cite{transformer}. Swin and PVT are recently proposed hierarchical vision Transformers (HVTs) which achieve promising results on various vision tasks. Moreover, as recent models usually have multiple variants in terms of the model depth and width, we denote them as ``Model-Ti/S/B'' to represent their tiny, small and base 
settings.

\noindent\textbf{Implementation details.}
By default, all models are trained on 8 V100 GPUs with a total batch size of 1,024 (128 per GPU) on ImageNet. We adopt AdamW~\cite{adamw} optimizer with a cosine decay learning rate scheduler. We set the initial learning rate and weight decay as $1\times10^{-3}$ and $5\times10^{-2}$, respectively. Furthermore, we adopt the same training strategies when comparing to each baseline. All experiments have adopted automatic mixed-precision training~\cite{low_pre_train} (also called FP16 or half-precision training) as it is widely used in recent work to accelerate the training process. The $\lambda$ in Eqs.~(\ref{eq:ema_alpha}) and (\ref{eq:ema_beta}) is set to 0.9, which is determined by a simple grid search on CIFAR-100. 
$\alpha$ and $\beta$ are initialised by the min-max values from the activations at the first training iteration. 
In Section~\ref{sec:ablation}, we conduct experiments of using different $\lambda$ to train DeiT-Ti on CIFAR-100 and provide visualisations for the evolution of quantization parameters.

\begin{table*}[]
\centering
\caption{Performance comparisons on DeiT-Ti with Mesa by using per-sample statistics (PS) and running estimates (RE). We report the Top-1 accuracy on ImageNet and CIFAR-100. ``*'' denotes our retrained baseline. Both experiments adopt the PyTorch implementation of Mesa.}
\renewcommand\arraystretch{1.2}
\label{tab:stat_ema}
\begin{tabular}{l|cccc}
Method &
  Train Memory (MB) &
  Train Throughput (images/s) &
  ImageNet Top-1(\%) &
  CIFAR-100 Top-1(\%) \\ \shline
DeiT-Ti*~\cite{deit}      & 4,149 & 1,196 & 71.9 & 64.8 \\
+ Mesa w/ PS & 2,117 & 372  & 71.9 & 65.1 \\
+ Mesa w/ RE & 2,000 & 431  & 72.1 & 65.2
\end{tabular}
\end{table*}

\begin{table*}[!htp]
\centering
\renewcommand\arraystretch{1.2}
\caption{Performance comparisons on DeiT with Mesa under stochastic rounding (SR) and nearest rounding (NR). We report the Top-1 accuracy on CIFAR-100 and ImageNet.}
\label{tab:stochastic_round}
\scalebox{1.0}{
\begin{tabular}{l|cccc}
Method &
  Train Memory (MB) &
  Train Throughput (images/s) &
  ImageNet Top-1(\%) &
  CIFAR-100 Top-1(\%) \\ \shline
DeiT-Ti~\cite{deit} & 4,149 & 1,196 & 71.9 & 64.8 \\
+ Mesa w/ NR                         & 1,855 & 635  & failed & 64.7 \\
+ Mesa w/ SR                          & 1,855 & 586  & 72.1 & 65.2
\end{tabular}
}
\end{table*}

\noindent\textbf{Results.}
In Table~\ref{tab:classification}, we report the ImageNet classification results of training DeiT and recent HVTs with Mesa. In general, Mesa can reduce around half of the memory consumption at training time while achieving comparable or even better performance than the strong baselines. For example, on DeiT-B and Swin-Ti, Mesa achieves the same performance as the baselines while reducing the memory footprint by 51\% and 55\%~\footnote{AMP~\cite{low_pre_train} training stores mixed FP16 and FP32 activations.
With Mesa only Int8 activations are saved, and thus we can
achieve more than 2× memory reduction.}, respectively. Remarkably, DeiT-Ti/S with Mesa even outperform the baselines by 0.2\% in the Top-1 accuracy. Our conjecture is that the approximated activations help to regularize the stochastic gradients when training Transformers, which therefore improves the model performance. Apart from this, we also visualize the training curves of DeiT-B with Mesa in Figure~\ref{fig:deit_b_training_curve}. 
As it shows, all curves under Mesa are consistent with those of baseline DeiT-B or even perform slightly better. For PVT-Ti, we observe a slight performance drop of 0.2\% in the Top-1 accuracy.
Our speculation is that PVTs are quite sensitive to train as the authors find that a deeper PVT even cannot converge with the same settings of PVT-Ti/S~\footnote{\url{https://github.com/whai362/PVT/issues/2}}.
In practice, the impact of Mesa on the model performance can be architecture-dependent. In general, Mesa performs consistently well on representative models. As shown in Table~\ref{tab:classification}, the effect on the performance is only around ±0.2\% for the compared baselines.

Besides, we notice that Mesa slows down the training speed.
As discussed in Section~\ref{sec:discussion}, the space-time trade-off is a common practice in the literature. In Table~\ref{tab:classification}, the experiments with Mesa compress almost all operations to achieve significant memory reduction, thus slowing down the throughput the most. However, in practice, one can determine the most memory-hungry operations with Mesa to achieve the ideal memory-saving target with acceptable training speed. We will show in Table~\ref{tab:compress_layer_types} and Table~\ref{tab:larger_model} that Mesa is able to compress different operations to achieve flexible memory-savings.
Furthermore, the training overhead can be largely offset by the data loading and communication cost in distributed training. For example, the training time of PVT-Ti with Mesa on ImageNet is only 15\% longer (520 vs. 600 GPU hours).
However, we also notice that the total GPU hours for training DeiT-B and Swin-Ti with Mesa are almost doubled. This suggests that training speed reduction varies among different architectures and model sizes, while the worst case may double the training time.

\vspace{-15pt}
\subsection{Ablation Studies} \label{sec:ablation}
\vspace{-5pt}
In this section, we provide ablation studies of Mesa. By default, we use a CUDA implementation of Mesa for the following experiments. The throughput and memory consumption at training time are measured with a batch size of 128 and an image resolution of 224$\times$224 on a single NVIDIA RTX 3090 GPU. Unless otherwise specified, we adopt the same training strategy as in Section~\ref{sec:main_res} for ImageNet experiments. On CIFAR-100, we train models with a total batch size of 256 on 2 GPUs and keep all other experiment settings as same as in Section~\ref{sec:main_res} except that the initial learning rate is linearly scaled to $2.5\times10^{-4}$.

\subsubsection{Per-sample Statistics vs. Running Estimates for Updating Quantization Parameters}
Previous ACT frameworks such as TinyScript~\cite{tinyscript} and ActNN~\cite{actnn} use per-sample statistics (PS) to calculate the quantization parameters. 
As discussed in~\ref{sec:head_wise_quant}, such approach can result in more additional memory consumption and computational cost during training.
In Table~\ref{tab:stat_ema}, we compare the approach of using per-sample statistics with using running estimates (RE) on ImageNet and CIFAR-100.
From the results, we observe that while both methods perform favourably against the baseline in terms of the Top-1 accuracy, the strategy of using running estimates achieves better performance, less memory consumption and faster throughput than using per-sample statistics. Note that both of the PS/RE experiments in Table~\ref{tab:stat_ema} adopt a PyTorch implementation of Mesa, thus the memory consumption and throughput at training time are slightly different from that of CUDA implementation. 
Besides, although the throughput on a single GPU is reduced by using Mesa, the total training time on ImageNet is similar to that of baseline DeiT-Ti as the communication overheads have more impact on the training speed under the distributed training scheme.

\subsubsection{Stochastic Rounding vs. Nearest Rounding} \label{sec:sr_nr}
To explore the effect of stochastic rounding in Mesa, we compare it with the commonly used nearest rounding by training DeiT-Ti with Mesa on CIFAR-100 and ImageNet. We report the results in Table~\ref{tab:stochastic_round}. As it shows, although training DeiT-Ti with nearest rounding can achieve competitive results on CIFAR-100, it fails to converge on ImageNet. Therefore, we speculate that stochastic rounding is important to guarantee good performance in Mesa. Also note that stochastic rounding does not increase the memory footprint during training. However, it is slightly slower than 
that of using nearest rounding, which attributes to the additional overhead from the implementation of stochastic rounding.

\begin{table}[]
\centering
\renewcommand\arraystretch{1.2}
\caption{Performance comparisons between symmetric quantization and asymmetric quantization on DeiT-Ti with Mesa. Both experiments adopt a PyTorch implementation of Mesa. We report the Top-1 accuracy on CIFAR-100. ``sym'' and ``asym'' denote symmetric and asymmetric quantization, respectively.}
\label{tab:aysm_sym}
\scalebox{0.85}{
\begin{tabular}{l|ccc}
Method &
  \begin{tabular}[c]{@{}c@{}}Train Memory\\ (MB)\end{tabular} &
  \begin{tabular}[c]{@{}c@{}}Train Throughput\\ (images/s)\end{tabular} &
  \begin{tabular}[c]{@{}c@{}}Top-1\\ (\%)\end{tabular} \\ \shline
DeiT-Ti~\cite{deit}            & 4,149 & 1,196 & 64.8 \\
+ Mesa w/ sym & 2,045 & 472  & 63.2 \\
+ Mesa w/ asym & 2,000 & 431  & 65.2
\end{tabular}
}
\end{table}

\begin{table}[]
\centering
\renewcommand\arraystretch{1.2}
\caption{Performance comparisons on DeiT-Ti with Mesa under different quantization granularities on CIFAR-100. ``Mesa w/ Layer'' means we train DeiT-Ti with Mesa under the layer-wise quantization. 
``Head'' indicates our proposed head-wise quantization}
\scalebox{0.85}{
\begin{tabular}{l|ccc}
Method &
  \begin{tabular}[c]{@{}c@{}}Train Memory\\ (MB)\end{tabular} &
  \begin{tabular}[c]{@{}c@{}}Train Throughput\\ (images/s)\end{tabular} &
  \begin{tabular}[c]{@{}c@{}}Top-1\\ (\%)\end{tabular} \\ \shline
DeiT-Ti~\cite{deit}     & 4,168 & 1,196 & 64.8    \\
+ Mesa w/ Layer   & 1,855 & 655  & 64.7 \\
+ Mesa w/ Head    & 1,855 & 586  & 65.2   
\end{tabular}
}
\label{tab:quant_granularity}
\end{table}

\begin{table}[]
\centering
\caption{Performance comparisons of different $\lambda$ based on DeiT-Ti. We report the Top-1 accuracy on CIFAR-100.}
\label{tab:effect_lambda}
\renewcommand\arraystretch{1.2}
\begin{tabular}{l|cc}
Method                   & $\lambda$ & Top-1 (\%)  \\ \shline
DeiT-Ti~\cite{deit}                  & -      & 64.8   \\ \hline
\multirow{4}{*}{+ Mesa} & 0.0    & 64.5   \\
                         & 0.9    & 65.2   \\
                         & 0.99   & 65.0   \\
                         & 0.999  & failed
\end{tabular}
\vspace{-10pt}
\end{table}

\subsubsection{Asymmetric Quantization vs. Symmetric Quantization} \label{sec:asym_sym}
Apart from the asymmetric quantization that used in Mesa, symmetric quantization is also widely adopted in the previous model quantization works~\cite{choi2018pact,lsq}. Under this scheme, the input $\bX$ is quantized by a scale factor $s = (2^b - 1) / \mathrm{max(|\bX|)}$ only, where $b$ is the bit width. In Table~\ref{tab:aysm_sym}, we compare the two quantization schemes based on CIFAR-100. From the results, we observe that although symmetric quantization achieves faster throughput during training, it does not surpass the baseline in terms of the Top-1 accuracy. The result is also consistent with a previous observation~\cite{ternary_bert} for quantizing BERT models. In fact, the activations in Transformers are skewed into negative values. We assume asymmetric quantization can provide a more tighter clipping range such that it helps to minimize the quantization error during training.

\begin{table*}[!htb]
\centering
\caption{Performance comparisons of compressing different operations based on DeiT-Ti with Mesa. We report the Top-1 accuracy on CIFAR-100.}
\renewcommand\arraystretch{1.2}
\label{tab:compress_layer_types}
\begin{tabular}{l|lcc}
Method    & Train Memory (MB) & Train Throughput (images/s) & Top-1 (\%) \\ \shline
DeiT-Ti~\cite{deit}  & 4,149              & 1,196                        & 64.8       \\
+ Mesa w/ MatMul        & 3,505 \fontsize{9pt}{9pt}\selectfont{(-15.5\%)}             & 729                         & 65.3       \\
+ Mesa w/ GELU      & 3,540 \fontsize{9pt}{9pt}\selectfont{(-14.7\%)}               & 1,031                        & 64.9       \\
+ Mesa w/ LayerNorm & 3,844 \fontsize{9pt}{9pt}\selectfont{(-7.4\%)}               & 1,059                        & 64.4       \\
+ Mesa w/ Softmax   & 3,485 \fontsize{9pt}{9pt}\selectfont{(-16.0\%)}             & 998                         & 64.8       \\
+ Mesa w/ All       & 1,855 \fontsize{9pt}{9pt}\selectfont{(-55.3\%)}               & 586                         & 65.2      
\end{tabular}
\vspace{-10pt}
\end{table*}

\begin{table*}[!htp]
\centering
\caption{Performance comparisons of compressing different modules based on DeiT-Ti with Mesa. We report the Top-1 accuracy on CIFAR-100.}
\renewcommand\arraystretch{1.2}
\label{tab:compress_msa_ffn}
\scalebox{1.0}{
\begin{tabular}{l|ccc}
Method   & Train Memory (MB) & Train Throughput (images/s) & Top-1 (\%)  \\ \shline
DeiT-Ti~\cite{deit}  & 4,149         & 1,196                        & 64.8    \\
+ Mesa w/ MSA      & 3,037 (-26.8\%)          & 772                         & 65.0 \\
+ Mesa w/ FFN      & 3,294 (-20.6\%)        & 888                         & 64.8 \\
+ Mesa w/ MSA + FFN      &  1,856 (-55.3\%)           & 597                         & 64.9   
\end{tabular}
}
\vspace{-10pt}
\end{table*}

\subsubsection{Effect of Different Quantization Granularities}
To explore the effect of different quantization granularities in Mesa, we train DeiT-Ti with Mesa and compare the proposed strategy to layer-wise quantization on CIFAR-100.
The results are shown in Table~\ref{tab:quant_granularity}.
Overall, benefiting from using running estimates in Mesa, all strategies consume a similar amount of memory during training as each layer only needs to save a few quantization parameters. 
With the layer-wise quantization, all activations at the same layer are quantized based on the same clipping range and offset, but it does not outperform the baseline.
On the other hand, the proposed head-wise quantization utilizes the bucketing strategies as in Figure~\ref{fig:bucketing} which distinguishes different statistics over different heads and channel groups. It achieves better performance than both the layer-wise strategy and the baseline, while making a good trade-off to the training speed. 
Additionally, we notice that grouping activations (except queries, keys, values and attentions) based on each hidden channel makes DeiT-Ti fail to converge on ImageNet. This suggests an appropriate quantization granularity is needed to stabilize Transformer training. 

\subsubsection{Effect of Different Decay Rates to Learn Quantization Parameters}\label{sec:eff_ema}
In Mesa, we utilise running estimates to learn the quantization parameters, which requires a decay rate $\lambda$ to tune. To understand the effect of different $\lambda$ in Mesa, we conduct experiments with DeiT-Ti on CIFAR-100 and report the results in Table~\ref{tab:effect_lambda}. Overall, we find that a suitable $\lambda$ is quite essential to help Mesa achieve good performance. 
Particularly, when $\lambda$ is 0, the quantization parameters only rely on the current batch statistics at each training iteration. However, such an approach cannot outperform the baseline.
Besides, we find a large $\lambda$ make DeiT-Ti fail to converge as it cannot timely adapt to the dynamic activation distributions during training. Overall, we find 0.9 achieves the best performance in practice, in which case we set $\lambda$ as 0.9 for all experiments by default.

\begin{table}[]
\centering
\caption{Comparisons of the largest models Mesa can train before
out-of-memory with the same batch size of 128 on one NVIDIA V100 32GB GPU. ``Depth'' refers to the model depth or the number of Transformer blocks. ``Width'' means the model width or the number of self-attention heads at each block. ``Resolution'' denotes the input image resolution during training.}
\renewcommand\arraystretch{1.2}
\label{tab:model_size}
\begin{tabular}{l|ccc}
Dim        & DeiT-B & DeiT-B w/ Mesa & Ratio \\ \shline
Depth      & 12     & 40             & \textbf{3.3}$\times$  \\
Width      & 12     & 26             & \textbf{2.2}$\times$ \\
Resolution & 224    & 336            & \textbf{1.5}$\times$ 
\end{tabular}
\end{table}

\subsubsection{Effect of Compressing Different Operations}
It is worth to mention that Mesa can compress different operations, or any combinations (\eg, GELU and LayerNorm) to achieve different memory-speed trade-offs. For example, a standard Transformer~\cite{vit,deit} consists of matrix multiplication (MatMul), GELU, LayerNorm and Softmax. The activations that are generated from these operations consume most of the GPU memory during training. In Table~\ref{tab:compress_layer_types}, we report the results of compressing different operations in DeiT-Ti based on CIFAR-100. In general, training with Mesa can achieve on par or even better performance than the baseline. At the same time, compressing different operations can achieve different memory-savings while also introducing different training overheads. When compressing all operations, we achieve the most memory-saving (around 50\%) while slowing down the training speed on a single GPU by half. In practice, \textit{Mesa allows researchers to flexibly set the trade-off according to their own target memory-saving and acceptable training overhead}. We show in Section~\ref{sec:train_larger_model} that by using different combinations of the compressed operations we can train larger models with a better memory-speed trade-off.

\subsubsection{Effect of Compressing Different Modules} \label{sec:eff_module}
MSA and FFN layers are the main modules of a Transformer model. Meanwhile, they consume most of the GPU memory at training time.
To study the effect of compressing different modules, we train DeiT-Ti with Mesa on CIFAR-100 and report the results in Table~\ref{tab:compress_msa_ffn}. Overall, training DeiT-Ti with Mesa achieves on par or better performance compared to the baseline. In particular, compressing the MSA or FFN layers in DeiT-Ti can reduce 27\% and 21\% memory footprint at training time, respectively. However, while compressing MSA layers can save more memory, it also results in slower throughput during training compared to compressing FFN layers. Finally, compressing MSA and FFN layers simultaneously brings the most memory savings, but it also leads to the slowest training speed. However, this overhead could be offset by the communication cost in distributed learning. In practice, one can also target different modules in a model to achieve an acceptable memory-speed trade-off.

\begin{table*}[]
\centering
\renewcommand\arraystretch{1.2}
\caption{Performance comparisons with larger batch size based on DeiT-Ti and Swin-S with Mesa. ``Total Memory'' indicates the total memory consumption over 8 GPUs. The GPU hours are calculated w.r.t. a single NVIDIA V100 GPU. We report the Top-1 accuracy on ImageNet.}
\scalebox{1.0}{
\begin{tabular}{l|cccc}
Method & Batch Size & Total Memory (GB) & GPU Hours & Top-1 (\%) \\ \shline
DeiT-Ti~\cite{deit}         & 1,024 & 33.4 & 440 & 71.9    \\
\textbf{DeiT-Ti w/ Mesa} & 2,048 & \textbf{28.8}    & 500 & \textbf{72.9} \\ \hline
Swin-S~\cite{swin}  & 1,024 & 150.7 & 968 & 83.0    \\ 
\textbf{Swin-S w/ Mesa} & 2,048 & \textbf{120.8}   & 1,160 & \textbf{83.1}
\end{tabular}
}
\label{tab:larger_bs}
\end{table*}

\subsubsection{Largest Models that Mesa can Train}
With the help of Mesa, we can re-invest the reduced memory by constructing a larger model or training with a larger image resolution. In Table~\ref{tab:model_size}, we report the largest models that Mesa can train before out-of-memory based on DeiT-B. Overall, Mesa is able to scale up the model depth by $3.3\times$ and width by $2.2\times$. Moreover, Mesa can train DeiT-B with $1.5\times$ larger image resolution. In practice, one can also decrease the expansion ratios in the FFN layers to scale up more on the depth/width/resolution. In Table~\ref{tab:model_size}, we adopt the default expansion ratio of four in DeiT-B. Furthermore, we will show in Section~\ref{sec:larger_batch_size} that Mesa enables us to train models with a larger batch size under the same memory budget.
\begin{figure*}[]
	\centering
	\includegraphics[width=0.75\linewidth]{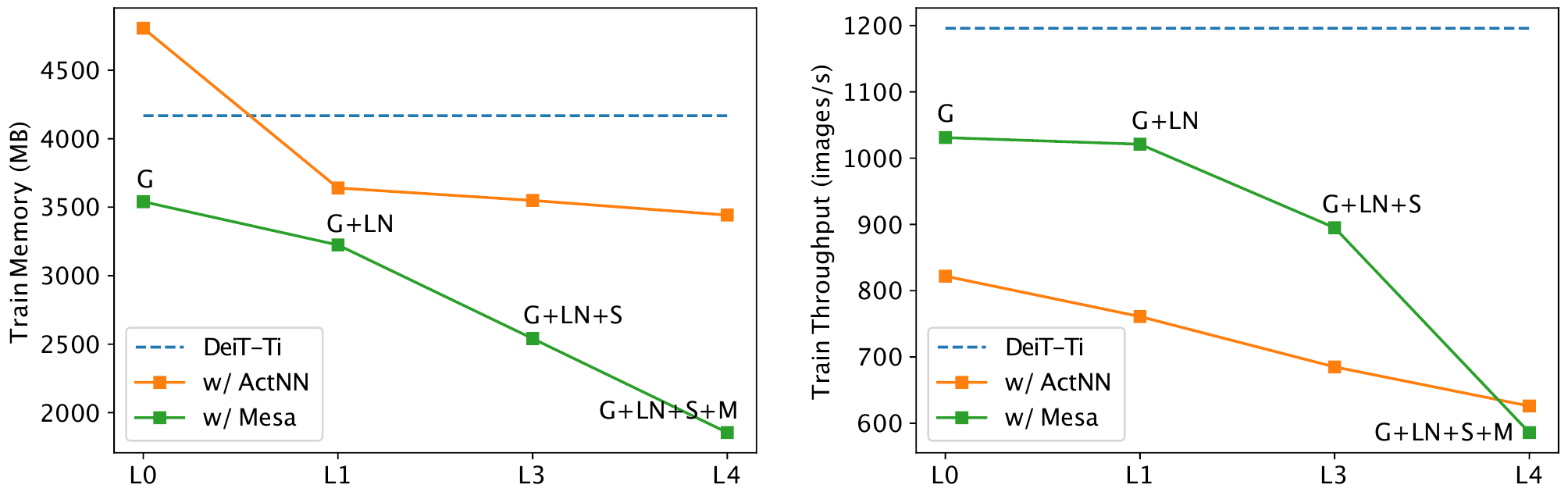}
	\caption{Memory consumption and throughput comparison at training time between ActNN and Mesa based on DeiT-Ti. ``L0'',``L1'',``L3'' and ``L4'' refer to different optimization levels in ActNN~\cite{actnn}. Note that L2 in ActNN achieves the same memory reduction as L1, and L5 does not further reduce memory cost. Therefore, we exclude L2 and L5 for brevity. We also denote ``G'', ``LN'', ``S'' and ``M'' as the GELU, LayerNorm, Softmax and MatMul operations, respectively. ``+'' means we simultaneously compress multiple operations with Mesa. For example, ``G+LN'' represents for compressing both the GELU and LayerNorm in DeiT. Compared to the different optimization levels in ActNN, Mesa can flexibly combine different operations and achieve better memory-speed trade-off. Best viewed in color.}
	\label{fig:compare_to_actnn}
\end{figure*}

\begin{table}[]
\centering
\caption{Sementic segmentation performance on ADE20K~\cite{ade20k} based on Semantic FPN~\cite{sem_fpn}. We use Swin-Ti as backbone and measure the performance by mIoU.}
\label{tab:sem_seg}
\renewcommand\arraystretch{1.2}
\scalebox{1.0}{
\begin{tabular}{l|cccc}
Backbone       & Batch Size & GPUs & mIoU (\%)  \\ \shline
Swin-Ti~\cite{swin}         & 16 & 8 & 41.5  \\
\textbf{Swin-Ti w/ Mesa} & 16 & 8 & \textbf{42.2}   \\
\textbf{Swin-Ti w/ Mesa} & 16 & 2 & \textbf{42.4}   \\
\end{tabular}
}
\end{table}

\begin{table}[]
\centering
\renewcommand\arraystretch{1.2}
\caption{Performance comparison between ActNN and the proposed Mesa based on DeiT-Ti. We adopt the default setting (L3, 2-bit) for ActNN. We report the Top-1 accuracy on CIFAR-100.}
\label{tab:actnn_mesa}
\scalebox{0.9}{
\begin{tabular}{l|ccc}
Method & \begin{tabular}[c]{@{}c@{}} Train Memory \\ (MB)\end{tabular}& \begin{tabular}[c]{@{}c@{}}Train Throughput\\ (images/s)\end{tabular}  & \begin{tabular}[c]{@{}c@{}}Top-1\\ (\%)\end{tabular} \\ \shline
DeiT-Ti~\cite{deit} & 4,149 & 1,196 & 64.8 \\
+ ActNN~\cite{actnn}   & 3,514 (-15.3\%) & 635 (-46.9\%)  & 64.9    \\
+ Mesa    & 1,855 (-55.3\%) & 586 (-51.0\%)  & 65.2   
\end{tabular}
}
\vspace{-10pt}
\end{table}

\subsubsection{Semantic Segmentation}
To explore the performance of Mesa on downstream tasks, we use Swin-Ti as backbone and Semantic FPN~\cite{sem_fpn} as the framework to evaluate the performance of semantic segmentation on ADE20K\cite{ade20k}. Following common practice~\cite{chu2021Twins}, we use AdamW optimizer with a poly learning rate schedule and train models with 80,000 iterations. 
We set the initial learning rate to $1\times10^{-4}$. All backbones are pretrained on the ImageNet dataset. 
As Table~\ref{tab:sem_seg} shows, training with Mesa outperforms the baseline by 0.7\% in mIoU. Furthermore, with the reduced memory footprint, we train the model with only two GPUs under the same setting while achieving 0.2\% more gains in mIoU, which again demonstrates the advantage of Mesa under the finite hardware resources.

\begin{table*}[!htb]
\centering
\caption{Experiments with Swin-B/L on CIFAR-100. Memory footprint is measured with a batch size of 64. ``G'', ``S'', ``LN'' denote GELU, Softmax, LayerNorm, respectively. ``all'' means we compress all types of operations.}
\label{tab:larger_model}
\begin{tabular}{l|ccccc}
Method         & Epochs  & Train Memory (MB) & Top-1 (\%) & Throughput (images/s) \\ \shline
Swin-B         & 300     & 13,717      & 78.0 & 111     \\
w/ Mesa all & 300     & 6,708 (-51\%)       & 78.0 & 55 (-50\%)    \\ 
w/ Mesa G+S & 300     & 9,714 (-29\%)       & 78.4 & 97 (-13\%)    \\ \hline
Swin-L         & 90      & 21,851      & 64.5  & 83    \\
w/ Mesa all & 90      & 11,708 (-46\%)    & 64.6   & 38 (-54\%)  \\  
w/ Mesa G+S+LN & 90      & 13,943 (-36\%)    & 65.0   & 67 (-19\%)       \\  
\end{tabular}
\end{table*}

\subsubsection{Effect of Larger Batch Size under the Same Memory Budget} \label{sec:larger_batch_size}
In our main results on ImageNet classification, we have shown that Mesa is able to reduce around half of the memory consumption at training time. More importantly, with Mesa, it is possible to train a Transformer with a larger batch size under the same memory budget. For example, training DeiT-Ti with the default strategy requires at least 34GB GPU memory under a total batch size of 1,024. However, with the same memory budget, we can double the batch size to better utilise available GPU cores and explore potential benefits of using larger batch size. Specifically, we train DeiT-Ti with Mesa using a total batch size of 2,048 on ImageNet with 8 NVIDIA V100 GPUs. As we can see from Table~\ref{tab:larger_bs}, Mesa still consumes less memory than that of mixed-precision training with a batch size of 1,024 (33.4GB vs. 28.8GB). Furthermore, it achieves 0.8\% gains in the Top-1 accuracy. Moreover, while the default training for Swin-S with a batch size of 2,048 will result in out of memory on 8 V100 GPUs, training with Mesa under the same batch size again consumes less memory and even achieves 0.1\% improvement in the Top-1 accuracy compared to the baseline. The overall performance indicates that Mesa enables us to train models with a larger batch size under the same memory budget, while still performing favourably against baselines.

\begin{table}[!htb]
\centering
\caption{Comparison with checkpointing (CP) and gradient accumulation (GA) based on Swin-Ti. Mesa is complementary to these techniques.}
\label{tab:compare_checkpoint_ga}
\scalebox{0.9}{
\begin{tabular}{l|ccc}
Method       & Batch Size & Memory (MB) & Throughput \\ \shline
Swin-Ti      & 128        & 11,812      & 356                \\
w/ Mesa      & 128        & 6,741 (-43\%)       & 276 (-22\%)                \\ \hline
w/ CP        & 128        & 6,885       & 303                \\
w/ Mesa + CP & 128        & 4,353 (-63\%)       & 244 (-31\%)                \\ \hline
w/ GA        & 64         & 6,209       & 333                \\
w/ Mesa + GA  & 64         & 3,682 (-69\%)       & 257 (-28\%)               
\end{tabular}
}
\end{table}

\subsubsection{Compared to ActNN} \label{sec:compare_actnn}
Existing ACT frameworks include BLPA~\cite{backprop_mem}, TinyScript\cite{tinyscript} and ActNN~\cite{actnn}. However, both BLPA and TinyScript target on specific CNN architectures (\eg, ResNet~\cite{resnet}), making it difficult for applying their methods into Transformers. Besides, ActNN only experiments with CNN-based models which ignores the heterogeneously distributed activations in MSA layers. In Table~\ref{tab:actnn_mesa}, we train DeiT-Ti with the default setting (L3, 2-bit) of ActNN on CIFAR-100. Compared to ActNN, Mesa covers more layers and achieves better memory-saving and performance with a slightly slower throughput. Furthermore, Mesa can achieve a better memory-speed trade-off than ActNN by targeting different operations. For example, in Figure~\ref{fig:compare_to_actnn}, we show the advantage of Mesa by comparing it with the different optimization levels in ActNN. 

\subsubsection{Effect of Training on Larger Models} \label{sec:train_larger_model}
To explore the effect of training larger models with Mesa, we train Swin-Base and Swin-Large with Mesa on CIFAR-100. Following the default setting in Swin~\cite{swin}, we train Swin-Large for 90 epochs. As shown in Table~\ref{tab:larger_model}, Mesa reduces half memory footprint while keeping their performance when compressing all operations. Moreover, since the throughput is controllable, we can compress the selected operations to achieve a better speed-memory trade-off with even slightly higher accuracy on CIFAR-100.

\begin{figure*}[]
	\centering
	\includegraphics[width=\linewidth]{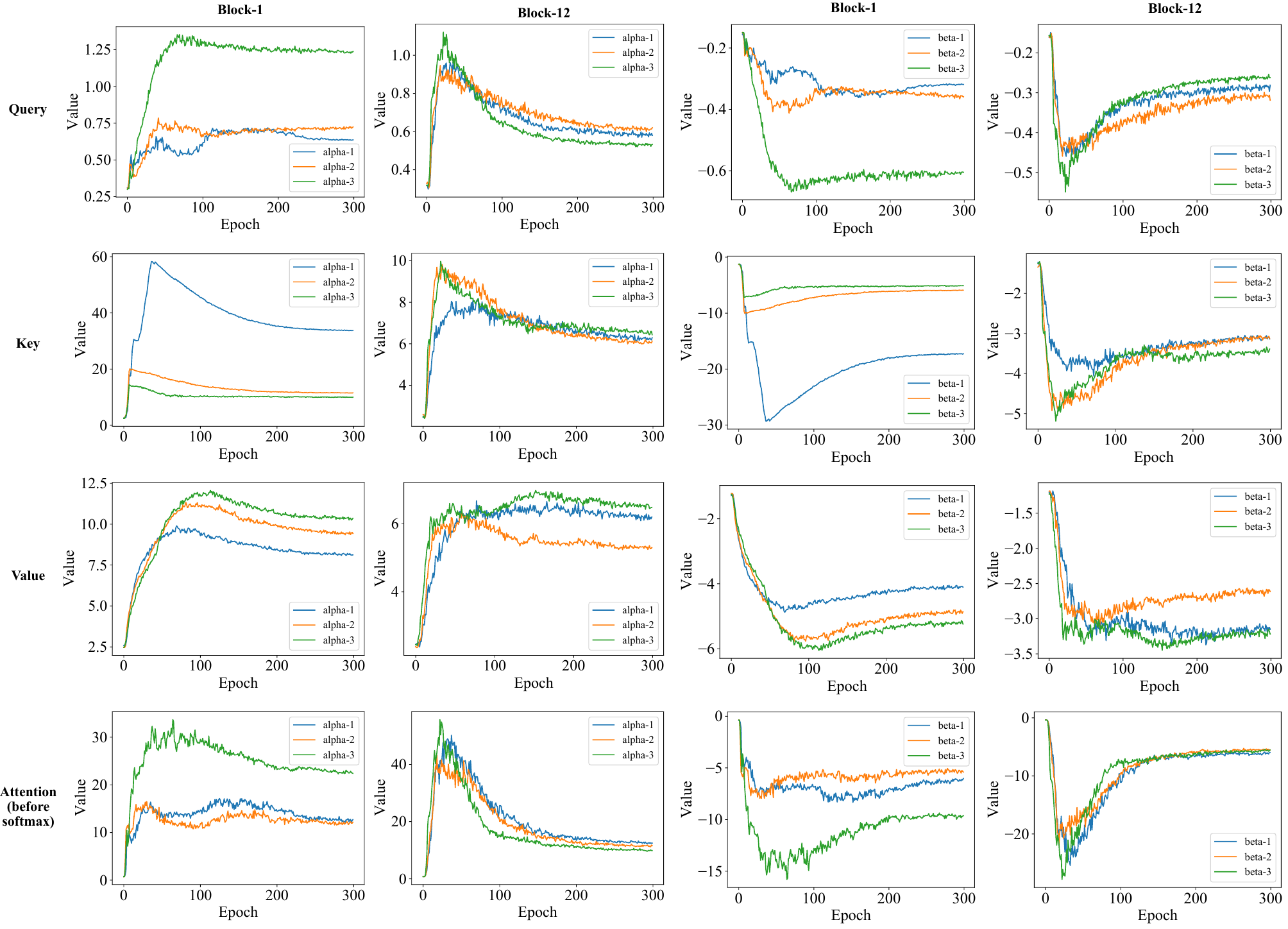}
	\caption{Evolution of  $\alpha$ and $\beta$ in the MSA layers of DeiT-Ti. We visualize the results for different blocks (columns) and activations (rows). Different colors represent different quantization groups. Best viewed in color.}
	\label{fig:quant_params_msa_alpha_beta}
\end{figure*}

\begin{figure*}[]
	\centering
	\includegraphics[width=\linewidth]{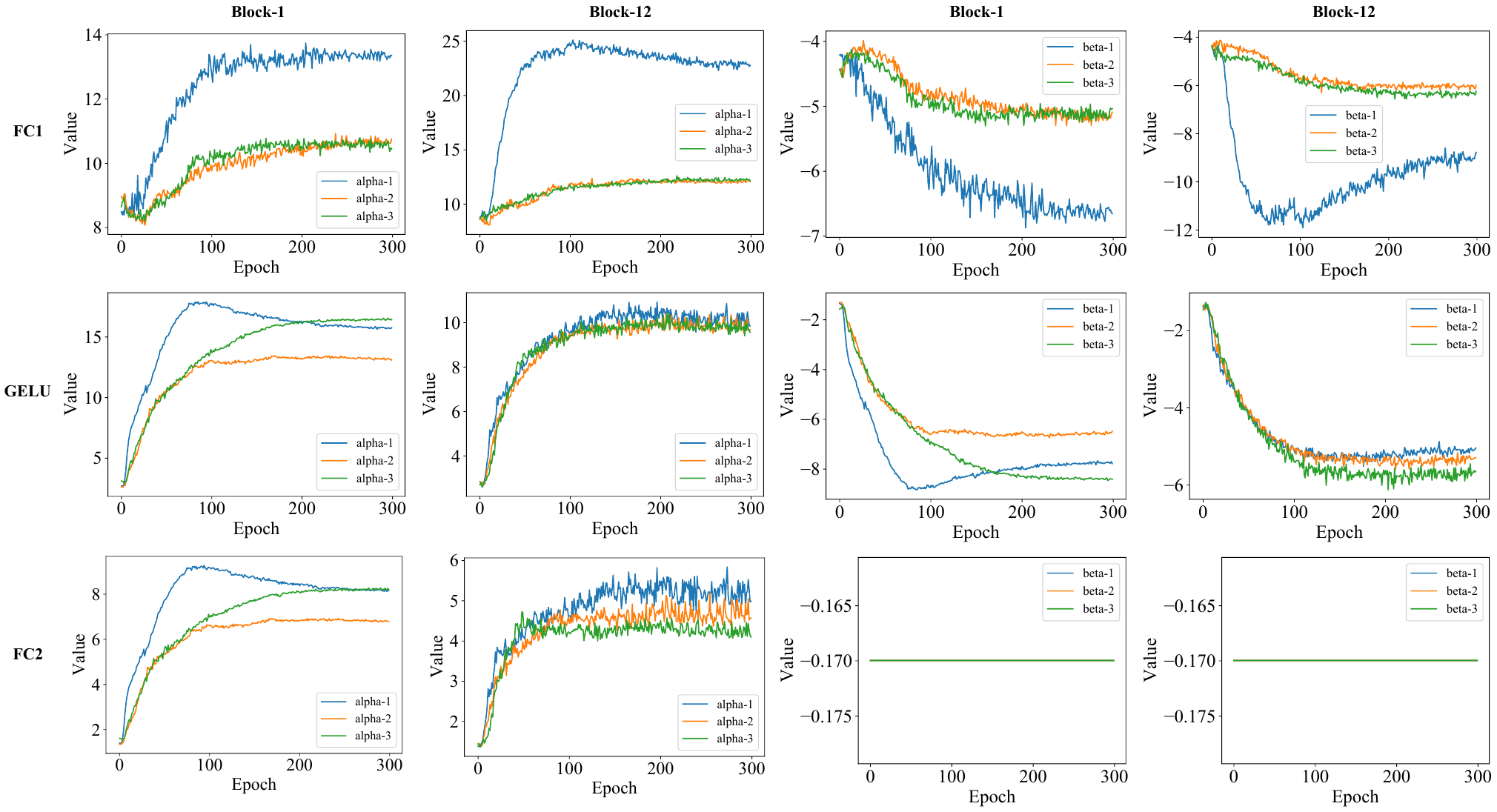}
	\caption{Evolution of $\alpha$ and $\beta$ in the FFN layers of DeiT-Ti.}
	\label{fig:quant_params_mlp_alpha_beta}
	\vspace{-20pt}
\end{figure*}

\subsubsection{Compared to Other Memory-saving Techniques} 
As discussed in Section~\ref{sec:discussion}, ACT frameworks including Mesa are complementary to checkpointing (CP), gradient accumulation (GA) and AMP. In fact, CP aims to store fewer tensors while ACT compresses saved tensors. Besides, GA saves memory by reducing the batch size, which makes it difficult for a fair comparison with related works. Moreover, even if most Transformers adopt LayerNorm instead of BatchNorm~\cite{batchnorm}, training ViTs such as DeiT can be sensitive to batch size. Therefore, reducing batch size can hurt model performance~\footnote{\url{https://github.com/facebookresearch/deit/issues/63}}. In contrast, Mesa saves memory \textit{without modifying the experimental hyperparameters}. Furthermore, Mesa is orthogonal to AMP. During training, Mesa uses exact activations during forward pass while storing Int8 activations for backpropagation to reduce memory consumption, while AMP quantizes weights, activations and gradients into FP16.

In Table~\ref{tab:compare_checkpoint_ga}, we compare Mesa with CP and GA. Overall, Mesa can save more memory while achieving comparable speed with CP. Note that CP has been highly optimized by PyTorch team, which is slightly not fair compared to our implementation. More importantly, Mesa can be combined with these techniques for more memory-saving due to complements.

\section{Visualization} \label{sec:evo_quant_param}
In this section,  we visualize the evolution of $\alpha$ and $\beta$ in DeiT-Ti during training on CIFAR-100. We set the number of quantization groups at each layer to 3.

\subsection{Evolution of $\alpha$ and $\beta$ in MSA Layers}
In Figure~\ref{fig:quant_params_msa_alpha_beta}, we show the evolution of $\alpha$ and $\beta$ in MSA layers during training, respectively.
In general, the quantization parameters at each head evolve differently, emphasizing the necessity of head-wise activation quantization in MSA layers. Besides, we find the quantization clipping range $\alpha$ increases significantly at the early stages, then becomes stable or deceases at the later stages. 
On the other hand, the quantization offset $\beta$ keeps decreasing at the beginning while tends to be stable or increases later during training. In particular, we find the quantization offsets are skewed to negative values, which indicates that the activations in MSA layers contain more negative values, except for the attentions after Softmax as they are non-negative values.

\vspace{-10pt}
\subsection{Evolution of $\alpha$ and $\beta$ in FFN Layers}
Figure~\ref{fig:quant_params_mlp_alpha_beta} shows the evolution of $\alpha$ and $\beta$ in FFN layers during training, respectively. Overall, the phenomenon is quite similar to that of MSA layers. In particular, we find the quantization offset at the second FC layer of FFN is always the minimum value (-0.17) of GELU. This indicates that $\beta$ can be fixed to -0.17 at this layer during training, which may help to reduce more training overhead.  We leave the space of further optimization for future work.

\section{Conclusion}
\label{conclusion}
In this work, we have presented Mesa, a memory-saving resource-efficient training framework for Transformers. Specifically, we save a low-precision approximated activations during training to achieve memory saving while using exact activations for the forward pass. The saved activations are used to calculated gradients during the backpropagation. Moreover, we identify the heterogeneous activation distributions in an MSA layer of a Transformer. For this, we proposed a head-wise activation quantization strategy, which groups the activations based on each self-attention head to capture better quantization clipping ranges and offsets. With Mesa, we can re-invest the reduced memory footprint by constructing a larger model or training with a larger batch size to explore potential benefits. Extensive experiments on ImageNet, CIFAR-100 and ADE20K have shown that Mesa can achieve flexible memory-savings during training while achieving comparable or even better performance than the default training strategy.


\bibliographystyle{IEEEtran}
\bibliography{reference}


\end{document}